\pdfoutput=1

\documentclass[11pt]{article}

\usepackage[]{acl}

\usepackage{times}
\usepackage{latexsym}

\usepackage[T1]{fontenc}

\usepackage[utf8]{inputenc}

\usepackage{microtype}

\usepackage{inconsolata}
\usepackage{booktabs} 
\usepackage{pifont}
\newcommand{\CHECK}{\textcolor{green}{\ding{51}}} 
\newcommand{\CROSS}{\textcolor{red}{\ding{55}}} 
\usepackage{graphicx} 
\usepackage{multirow}
\usepackage{multicol}
\usepackage{arydshln}
\usepackage{color}
\usepackage{amssymb}
\usepackage[T1]{fontenc}
\usepackage{amsmath}

\title{An Embarrassingly Simple Approach for LLM with Strong ASR Capacity}

\author{
Ziyang Ma\textsuperscript{$\spadesuit$}, 
Guanrou Yang\textsuperscript{$\spadesuit$},
Yifan Yang\textsuperscript{$\spadesuit$}, \\
\textbf{
Zhifu Gao\textsuperscript{$\heartsuit$},
Jiaming Wang\textsuperscript{$\heartsuit$},
Zhihao Du\textsuperscript{$\heartsuit$},
Fan Yu\textsuperscript{$\heartsuit$},
Qian Chen\textsuperscript{$\heartsuit$},
Siqi Zheng\textsuperscript{$\heartsuit$},} \\
\textbf{
Shiliang Zhang\textsuperscript{$\heartsuit$},
Xie Chen\textsuperscript{$\spadesuit$}\footnotemark[2],
} \\
$^\spadesuit$ MoE Key Lab of Artificial Intelligence, AI Institute,  \\
X-LANCE Lab, Shanghai Jiao Tong University, Shanghai, China  \\
$^\heartsuit$ Alibaba Group, China \\
}

\begin{document}
\maketitle

\renewcommand{\thefootnote}{\fnsymbol{footnote}}
\footnotetext[2]{Corresponding author}
\renewcommand{\thefootnote}{\arabic{footnote}}

\begin{abstract}
In this paper, we focus on solving one of the most important tasks in the field of speech processing, i.e., automatic speech recognition (ASR), with speech foundation encoders and large language models (LLM). 
Recent works have complex designs such as compressing the output temporally for the speech encoder, tackling modal alignment for the projector, and utilizing parameter-efficient fine-tuning for the LLM. 
We found that delicate designs are not necessary, while an embarrassingly simple composition of off-the-shelf speech encoder, LLM,  and the only trainable linear projector is competent for the ASR task. 
To be more specific, we benchmark and explore various combinations of LLMs and speech encoders, leading to the optimal LLM-based ASR system,  which we call SLAM-ASR\footnote{SLAM-ASR is a subproject of SLAM-LLM, where SLAM stands for \textbf{S}peech, \textbf{L}anguage, \textbf{A}udio and \textbf{M}usic. Working in progress and will open-source soon.}. 
The proposed SLAM-ASR provides a clean setup and little task-specific design, where only the linear projector is trained. 
To the best of our knowledge, SLAM-ASR achieves the best performance on the Librispeech benchmark among LLM-based ASR models and even outperforms the latest LLM-based audio-universal model trained on massive pair data. 
Finally, we explore the capability emergence of LLM-based ASR in the process of modal alignment. 
We hope that our study can facilitate the research on extending LLM with cross-modality capacity and shed light on the LLM-based ASR community. 

\end{abstract}

\vspace{-0.2cm}
\section{Introduction}
\vspace{-0.2cm}
Automatic speech recognition (ASR) stands as a cornerstone in the realm of intelligent speech technology, enabling machines to understand and transcribe human speech. The significance of ASR in enhancing human-computer interaction and accessibility makes it a crucial area of research and applications in the field of speech processing. 

The evolution of ASR technology has been marked by the adoption of various paradigms, each representing a leap forward in terms of accuracy, efficiency, and applicability~\cite{e2ereview}.
Among these, supervised methods including connectionist temporal classification (CTC)~\cite{CTC}, attention-based encoder-decoder (AED)~\cite{AED}, recurrent neural network transducer (RNN-T)~\cite{RNN-T} and their variants have been pivotal. 
In addition, employing self-supervised methods for pre-training followed by supervised methods for fine-tuning has also proven to be effective~\cite{baevski2020wav2vec, hsu2021hubert, chen2022wavlm, ma2022mt4ssl, yang2023fast}. 
However, each paradigm comes with its own set of challenges and limitations, such as the need for extensive labeled data, difficulties in capturing long-range context dependencies in speech, and huge training costs. 

In this evolving landscape, the advent of large language models (LLMs) has introduced a groundbreaking paradigm: Multimodal large language models (MLLMs) framework~\cite{liu2023visual, BLIP2, gao2023llama}, based on a decoder-only architecture. This innovative approach diverges from traditional ASR by utilizing the immense generative capacity of LLMs, which are pre-trained on vast corpora encompassing diverse linguistic contexts, leading to LLM-based ASR. 
The evolution of the ASR paradigm from previous NN-based ASR models to LLM-based ASR models, stresses differences across loss and criterion design, text prior knowledge, and model scale. 
This paradigm harnesses pre-existing linguistic knowledge, enabling a more holistic understanding of language, which in turn, translates to significant improvements in the speech recognition task. 

The architecture of LLM-based ASR can be conceptualized as consisting of three primary components: a speech encoder, a projector, and an LLM. 
Recent works in LLM-based ASR often venture into complex designs, such as compressing the output temporally from the speech encoder~\cite{wu2023decoder, fathullah2023prompting}, tackling modal alignment with the projector~\cite{tang2023salmonn, yu2023connecting}, and fine-tuning the LLM partly or fully~\cite{wu2023decoder, li2023prompting, tang2023salmonn, wang2023lauragpt}. 
Despite these efforts, the outcomes have not always met expectations, indicating a potential misalignment between the complexity of designs and the efficacy of real-world speech recognition tasks. This observation led to a pivotal realization in our research: the essence of an effective LLM-based ASR system lies in the synergy of a powerful speech encoder and a suitable LLM, and then, most notably, a single trainable linear projector is enough to align between modalities. Our findings challenge the prevailing notion that complexity equates to superiority in LLM-based ASR system design. 

In this work, we first benchmark the automatic speech recognition task performance with different combinations of well-known speech encoders and the latest released large language models. 
Experiments show that LLMs with supervised fine-tuning (SFT,  a.k.a. chat model) perform better than raw pre-trained LLMs for the ASR task, while speech encoders fine-tuned with limited data from self-supervised models outperform supervised foundation ASR encoders.
Building upon these insights, we propose SLAM-ASR, in which only a linear projector is trained to conduct the ASR task. SLAM-ASR only requires $4$ GPUs for $4$ hours of training to achieve state-of-the-art performance on the Librispeech~\cite{panayotov2015librispeech} corpus, compared with other LLM-based ASR models and a series of previous best performing NN-based ASR models. 
Besides, our work embarks on an in-depth exploration of the ability of LLM-based ASR models.  Interestingly,
 we observe the capability emergence phenomenon during LLM-based ASR training. 
The benchmark and experimental exploration show how we harvest the exciting result step by step with a clean setup and little task-specific design. 

\section{Speech Recognition Meets Large Language Model}

\begin{table}[htbp]
\centering
\caption{ASR Paradigm with representative models. \textbf{QF} means variants of Q-Former~\cite{BLIP2}. 
Both \textbf{QF} and \textbf{Linear} are projector modules used to align the speech encoder and the LLM. 
}
\label{tab:paradigm}
\resizebox{1\linewidth}{!}{
\begin{tabular}{lcc}
\hline
\textbf{Model} &\textbf{Loss} &\textbf{Learnable} \\
\hline
\multicolumn{3}{l}{\textbf{\textit{Previous NN-based ASR}}} \\
\hline
Quartznet~\cite{kriman2020quartznet} & CTC & All \\
Whisper~\cite{radford2023robust} & AED & All\\
Branchformer~\cite{peng2022branchformer} & CTC + AED & All\\
Conformer~\cite{gulati2020conformer} & RNN-T & All\\
Zipformer~\cite{yao2023zipformer} & Pruned RNN-T & All \\
Paraformer~\cite{gao2022paraformer} & CIF & All \\
\hline
\multicolumn{3}{l}{\textbf{\textit{LLM-based ASR}}} \\
\hline
LauraGPT~\cite{wang2023lauragpt} &\multirow{9}{1cm}{Decoder-Only, \\ Cross Entropy}  & All \\
SpeechGPT~\cite{zhang2023speechgpt} &  & LLM \\
\citeposs{li2023prompting} & & Encoder, LLM Adapter \\
SpeechLLaMA~\cite{wu2023decoder} &  & Encoder, LLM LoRA \\
Qwen-Audio~\cite{chu2023qwen} &  & Encoder, Linear \\
SALMONN~\cite{tang2023salmonn} &  & QF, LLM LoRA \\
\citeposs{fathullah2023prompting} &  & Linear, LLM LoRA \\
\citeposs{yu2023connecting}  &  & QF \\
\textbf{SLAM-ASR} &  & \textbf{Linear} \\
\hline
\end{tabular}
}
\end{table}

\subsection{Previous NN-based ASR}
Previous NN-based ASR systems are designed to align the speech signal with the label sequence accurately. 
As shown in table~\ref{tab:paradigm}, different paradigms are carried out with a series of representative models. 
Quartznet~\cite{kriman2020quartznet} leverages CTC~\cite{CTC}, the first E2E technology widely adopted in ASR, yet facing performance limitations due to its frame-independent assumption. 
Whisper~\cite{radford2023robust} utilizes massive pair speech-text data to train the attention-based encoder-decoder~\cite{AED} (AED, a.k.a. LAS in ASR) architecture, empowering the model with the ability to recognize and translate speech in multiple languages. 
Branchformer~\cite{peng2022branchformer} employs a hybrid architecture that combines CTC and AED~\cite{AED}, the integration of the attention mechanism addresses this limitation by introducing implicit language modeling across speech frames.
Conformer~\cite{gulati2020conformer} utilizes neural transducer~\cite{RNN-T}, which directly discards the frame-independent assumption by incorporating a label decoder and a joint network, resulting in superior performance.
Zipformer~\cite{yao2023zipformer} adopts Pruned RNN-T~\cite{PrunedRNN-T}, which is a memory-efficient variant of the transducer loss, utilizing the pruned paths with minor posterior probabilities.
Paraformer~\cite{gao2022paraformer} uses Continuous Integrate-and-Fire (CIF)~\cite{CIF}, which offers a soft and monotonic alignment mechanism, estimating the number of tokens and generating hidden variables.

\subsection{Existing LLM-based ASR}
LLM-based ASR models adopt decoder-only architectures based on a pre-trained LLM as a new paradigm.
LauraGPT~\cite{wang2023lauragpt} connects a modified Conformer~\cite{gulati2020conformer} encoder with Qwen-2B~\cite{bai2023qwen} for end-to-end training for multiple speech and audio tasks, with full parameter fine-tuning performed. 
SpeechGPT~\cite{zhang2023speechgpt} discretizes speech tokens with HuBERT~\cite{hsu2021hubert} and fine-tunes the LLaMA-13B~\cite{touvron2023llama1} with multiple stages. 
Although both models are computationally expensive, their performance is limited. 
\cite{li2023prompting} and \cite{wu2023decoder} propose to use inserted Gated-XATT-FFN~\cite{alayrac2022flamingo} or side-branched LoRA~\cite{hu2021lora} to fine-tune the LLM partially for conducting ASR task, along with a trainable speech encoder. 
Qwen-Audio~\cite{chu2023qwen} is an audio-universal model, which uses massive pair data to fine-tune the encoder initialized from the Whisper-large~\cite{radford2023robust} model, optimized using the loss of the frozen Qwen-7B~\cite{bai2023qwen} output for backpropagation. 
All these models require finetuning the encoder. 
SALMONN~\cite{tang2023salmonn} uses Whisper-large~\cite{radford2023robust} and BEATs~\cite{chen2022beats} to encode speech and audio, respectively, along with a window-level Q-Former (win-QF), can perform a variety of audio tasks. 
\cite{fathullah2023prompting} connects Conformer with LLaMA-7B to successfully conduct monolingual and multilingual ASR. 
These models require the use of LoRA to be effective. 
The most intimate work is ~\cite{yu2023connecting}, which achieves good results on ASR using only segment-level Q-Former (sef-QF) similar to win-QF as the projector. 
The random concatenation training strategy is designed to alleviate the natural problem of Whisper~\cite{radford2023robust} requiring an input speech of $30$ seconds. 

\subsection{Proposed Method}
\begin{figure}[t]
    \centering
    \includegraphics[width=1\linewidth]{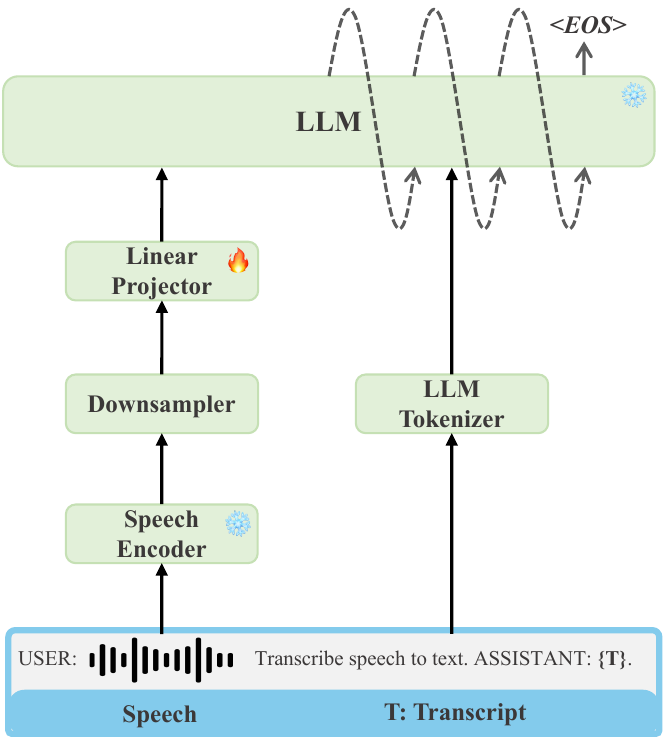}
    \caption{A brief pipeline of SLAM-ASR, at the core of which is a frozen speech encoder and a frozen LLM, with the only trainable linear projector to align between speech and text modalities. }
    \label{fig:model}
\end{figure}

As shown in Figure~\ref{fig:model}, an embarrassingly simple framework is proposed to train the SLAM-ASR model. 
For each sample, given speech $\mathbf{X^S}$, the corresponding transcript $\mathbf{X^T}$, and the prompt $\mathbf{X^P}$, we first convert the speech into speech features through the speech encoder, which can be written as:
\begin{equation}
\mathbf{H^S} = Encoder(\mathbf{X^S}), 
\end{equation}
where $\mathbf{H^S} = [h^S_1, \cdots, h^S_{T}]$ has $T$ frames in the temporal dimension. 
Due to the sparsity of speech representation, the speech features sequence $\mathbf{H^S}$ is still very long for the LLM to tackle\footnote{Speech features are 25, 50, or 100 frames per second in general. }, we downsample the speech with a downsampler. 
More explicitly, we concatenate every $k$ consecutive frames in the feature dimension to perform a $k$ times downsampling, leading to $\mathbf{Z^S} = [z^S_1, \cdots, z^S_{N}]$, where
\begin{equation}
z^S_i = h^S_{k*i} \oplus h^S_{k*i+1} \oplus \cdots \oplus h^S_{k*i+k-1},
\end{equation}
and
\begin{equation}
N = T // k. 
\end{equation}
Next, a projector is applied to transform the speech features $\mathbf{Z^S}$ into $\mathbf{E^S}$ with the same dimension as the LLM input embedding. 
In our experiments, we use a single hidden layer followed by a ReLU activation and a regression layer as the projector, donated as: 
\begin{equation}
    \mathbf{E^S} = Linear ( ReLU ( Linear ( \mathbf{Z^S} ) ) ).
\end{equation}
Finally, we feed the speech embedding $\mathbf{E^S}$, transcript embedding $\mathbf{E^T}$, and  prompt embedding $\mathbf{E^P}$ into the template to compose the final input $\mathbf{E}$ of LLM, donated as: 
\begin{equation}
\mathbf{E^T} = Tokenizer(\mathbf{X^T}), 
\end{equation}
\begin{equation}
\mathbf{E^P} = Tokenizer(\mathbf{X^P}), 
\end{equation}
\begin{equation}
\mathbf{E} = \begin{cases}
Template(\mathbf{E^S}, \mathbf{E^P}, \mathbf{E^T}) & \text{if training}, \\
Template(\mathbf{E^S}, \mathbf{E^P}) & \text{if inference},
\end{cases}
\end{equation}
wherein the template is detailed in Section~\ref{sec:training} and Section~\ref{sec:inference}. 

\section{Experiment Setup}
Our experimental procedure obeys the KISS \textit{(\textbf{K}eep \textbf{I}t \textbf{S}imple, \textbf{S}tupid!)} principle to investigate the most critical factors for LLM-based ASR. 

\subsection{Models and Modules}
\subsubsection{Speech Encoder}
Two types of speech encoders are investigated in this paper, which are supervised speech encoders trained on massive speech-text pair data and self-supervised speech encoders trained on large-scale unlabeled speech data. 
For supervised foundation models, we mainly survey the well-known Whisper~\cite{radford2023robust} family of models\footnote{\url{https://github.com/openai/whisper}} ranging from tiny to large, including \textit{whisper-tiny, whisper-base, whisper-small, whisper-medium} and \textit{whisper-large-v2}. 
We discard the decoder of each Whisper model and only use the encoder as a feature extractor. 
We also investigate \textit{Qwen-Audio Encoder}\footnote{\url{https://github.com/QwenLM/Qwen-Audio}}, the encoder fine-tuned from \textit{whisper-large-v2} checkpoint on large-scale speech, audio and music data, released along with Qwen-Audio~\cite{chu2023qwen} model. 
For self-supervised models, we investigate \textit{HuBERT}\footnote{\url{https://github.com/facebookresearch/fairseq/tree/main/examples/hubert}} and \textit{WavLM}\footnote{\url{https://github.com/microsoft/unilm/tree/master/unilm}} in different scales, either raw pre-trained or further fine-tuned. 
For the base-size models, both HuBERT~\cite{hsu2021hubert} and WavLM~\cite{chen2022wavlm} perform self-supervised pre-training on LibriSpeech~\cite{panayotov2015librispeech} corpus with $960$ hours. For the large-size models, HuBERT is trained on LibriLight~\cite{kahn2020libri} corpus with $60,000$ hours, while WavLM is trained on the much larger $94,000$ hours data including LibriLight~\cite{kahn2020libri}, VoxPopuli~\cite{wang2021voxpopuli}, and GigaSpeech~\cite{chen2021gigaspeech}. Furthermore, HuBERT provides pre-trained models of X-Large size, which is the largest publicly available self-supervised speech encoder. 
All the models mentioned in this section are obtained from their official repositories. 
Refer to Section~\ref{sec:encoder} for details of the parameters and hidden size of each specific model. 

\subsubsection{LLM}
Two types of large language models are investigated in this paper, which are raw pre-trained LLMs without supervised fine-tuning and chat LLMs with SFT (along with RLHF if conducted). 
For the pre-trained LLMs, we try \textit{TinyLLaMA}~\cite{zhang2024tinyllama}\footnote{\url{https://huggingface.co/TinyLlama/TinyLlama-1.1B-Chat-v0.4}} of the 1B-magnitude and \textit{LLaMA-2}~\cite{touvron2023llama2}\footnote{\url{https://huggingface.co/meta-llama/Llama-2-7b-hf}} of the 7B-magnitude.
For the chat LLMs, \textit{TinyLLaMA-Chat}\footnote{\url{https://huggingface.co/TinyLlama/TinyLlama-1.1B-intermediate-step-1431k-3T}} of the 1B-magnitude, \textit{Phi-2}\footnote{\url{https://huggingface.co/microsoft/phi-2}} of the 2B-magnitude, \textit{LLaMA-2-Chat}\footnote{\url{https://huggingface.co/meta-llama/Llama-2-7b-chat-hf}} and \textit{Vicuna}~\cite{chiang2023vicuna}\footnote{\url{https://huggingface.co/lmsys/vicuna-7b-v1.5}} of the 7B-magnitude are considered. 
Refer to Section~\ref{sec:llm} for details of the parameters and hidden size of each specific LLM. 

\subsubsection{Projector}
The projector can be viewed as an adaptor for other modalities to perform alignment with LLM. 
In all our experiments, the output of the speech encoder is $50$ Hz, and the downsampling rate $k=5$, leading to the input speech features $\mathbf{E^S}$ of the large model being $10$ Hz. 
The hidden layer dimension is set to $2048$, while the dimension of the speech encoder output $\mathbf{H^S}$ and the LLM input dimension vary depending on the model used, respectively.

\subsection{Dataset}
To evaluate the capabilities of the LLM-based ASR models, we use the most widely used benchmark for the ASR task, the standard Librispeech benchmark with $960$ hours of training data without any data augmentation or splicing. 
We use the dev-other subset as the validation set and test-clean/test-other as the test sets, each of which contains $10$ hours of speech. 

\subsection{Training Detail}
\label{sec:training}
During training, the data is organized in the following format:
``\textit{USER: \textless S\textgreater\ \textless P\textgreater\ ASSISTANT:\ \textless T\textgreater}'',
where \textit{\textless S\textgreater} represents speech embedding, \textit{\textless P\textgreater} represents the prompt, and \textit{\textless T\textgreater} represents the corresponding transcribed text. We only compute the loss on \textit{\textless T\textgreater}, as is common practice.
For the optimizing strategy, we use AdamW~\cite{loshchilov2017decoupled} with a max learning rate of $1 \times 10^{-4}$ without a weight decay. 
For the learning rate scheduler, we conduct warmup at the first $1,000$ steps and then keep the maximum learning rate for training all the time.
The max training step is set to $100,000$, but we will stop early if the loss on the validation set does not decrease. 
For the audio embedding provided by the Whisper family of models, we found that not padding would affect the performance. As a result, we pad the speech to $30$ seconds for all Whisper models and the batch size is set to $4$. For other models, the length of the input audio remains consistent with the original length in the temporal dimension, and the batch is set to $6$, which greatly improves the efficiency of training and inference, compared to Whisper models. 

\subsection{Inference Detail}
\label{sec:inference}
During inference, the data is organized in the following format:
``\textit{USER: \textless S\textgreater\ \textless P\textgreater\ ASSISTANT:}'',
where large language models answer autoregressively.  
Typically, LLMs utilize sampling algorithms to generate diverse textual outputs. 
Since speech recognition is a sequence-to-sequence task with deterministic outputs, we use beam search with $beam=4$ to output the hypothesis corresponding to the speech. 

\section{Exploration}
In this section, we first give \textbf{a basic benchmark} of combinations of different LLMs and speech encoders and find that chat models perform better than raw pre-trained LLMs on the ASR task. 
We next benchmark different chat models and find Vicuna to be \textbf{a suitable LLM} and fine-tuned HuBERT to be \textbf{a powerful speech encoder} for conducting the ASR task. 
Finally, we propose \textbf{SLAM-ASR}, and compare SLAM-ASR with state-of-the-art previous NN-based ASR models and the latest best-performing LLM-based ASR models. 

\subsection{A Basic Benchmark}
\label{sec:benchmark}
To begin with, we benchmark Whisper models with different sizes on pre-trained LLMs and supervised fine-tuned LLMs. 
We pick TinyLLaMA of the 1B-magnitude and LLaMA-2 of the 7B-magnitude to make a preliminary assessment. 
As shown in Table~\ref{tab:benchmark}, the performance of the ASR task improves as the speech encoder parameter size increases, but the improvement is of diminishing marginal benefit for the Whisper family of models. 
For the choice of LLMs, the chat models work better than the pre-trained models, regardless of the size. 
One possible explanation is that the chat models take speech embedding as a form of ``language'' and perform a machine translation task, which is activated during the SFT process. 

\begin{table*}[htbp]
\centering
\caption{A base benchmark with different combinations of speech encoders and LLMs to conduct LLM-based ASR. We benchmark Whisper models with different sizes on pre-trained models and chat models with different scales.}
\label{tab:benchmark}
\resizebox{1\linewidth}{!}{
\begin{tabular}{l|cccc|cccc}
\hline
\multirow{3}{*}{\textbf{Speech Encoder}} & \multicolumn{4}{c|}{\textbf{Pre-trained Model}} & \multicolumn{4}{c}{\textbf{Chat Model}} \\
& \multicolumn{2}{c}{TinyLLaMA} & \multicolumn{2}{c|}{LLaMA-2} & \multicolumn{2}{c}{TinyLLaMA-Chat} & \multicolumn{2}{c}{LLaMA-2-Chat} \\
& test-clean & test-other & test-clean & test-other & test-clean & test-other & test-clean & test-other \\
\hline
Whisper-tiny &12.72 &21.64 &16.16 &25.17 &9.55 &21.01 &8.97 &18.77 \\
Whisper-base &7.35 &15.89 &17.46 &21.84 &7.03 &15.92 &6.37 &12.98 \\
Whisper-small &6.61 &11.81 &6.41 &10.88 &5.94 &11.5 &4.51 &8.94 \\
Whisper-medium &4.65 &8.95 &3.35 &6.10 &5.01 &8.67 &2.71 &6.37 \\
Whisper-large &4.39 &8.22 &3.01 &7.15 &4.33 &8.62 &2.72 &6.79 \\
\hline
\end{tabular}
}
\end{table*}

\subsection{Exploration in LLMs}
\label{sec:llm}
Next, we fix the speech encoder as Whisper-large and then explore a better large language model. 
As shown in Table~\ref{tab:benchmark_llm}, the Phi-2 chat model with 2.78B parameters has a comparable word error rate with LLaMA-2 with 6.74B parameters on test-other. 
Vicuna is an open-source chat LLM fine-tuned on user-shared conversational data collected from ShareGPT\footnote{\url{https://sharegpt.com}}, utilizing LLaMA as a pre-trained LLM. 
The LLM-based ASR model shows better results when Vicuna is used as the LLM compared with LLaMA-2 and LLaMA-2-Chat. 
All the above experimental results confirm the capability of chat models on LLM-based ASR systems. 

\begin{table*}[htbp]
\centering
\caption{Explore the performance with different LLMs for LLM-based ASR. The projector is fixed with linear layers and the speech encoder is fixed with Whisper-large-v2.}
\label{tab:benchmark_llm}
\resizebox{1\linewidth}{!}{
\begin{tabular}{lccccc}
\hline
\multirow{2}{*}{\textbf{LLM}} & \multirow{2}{*}{\textbf{\#LLM Params}} & \multirow{2}{*}{\textbf{Hidden Size}} & \multirow{2}{*}{\textbf{\#Projector Params}} & \multicolumn{2}{c}{\textbf{WER(\%) $\downarrow$}} \\
& & & & \textbf{test-clean} & \textbf{test-other} \\
\hline
\multicolumn{6}{l}{\textbf{\textit{Pre-trained Model}}} \\
\hline
TinyLLaMA & 1.10B & 2048 & 17.31M & 4.39 & 8.22 \\
LLaMA-2 & 6.74B & 4096 & 21.50M & 3.01 & 7.15 \\
\hline
\multicolumn{6}{l}{\textbf{\textit{Chat Model}}} \\
\hline
TinyLLaMA-Chat & 1.10B & 2048 & 17.31M & 4.33 & 8.62 \\
Phi-2 & 2.78B & 2560 & 18.35M & 3.88 & 7.19 \\
LLaMA-2-Chat & 6.74B & 4096 & 21.50M & 2.72 & 6.79 \\
Vicuna & 6.74B & 4096 & 21.50M & 2.58 & 6.47 \\
\hline
\end{tabular}
}
\end{table*}

\begin{table*}[htbp]
\centering
\caption{Explore the performance with different speech encoders for LLM-based ASR. The projector is fixed with linear layers and LLM is fixed with Vicuna-7B-v1.5. LS-960 means the Librispeech 960 hours dataset. }
\label{tab:benchmark_encoder}
\resizebox{1\linewidth}{!}{
\begin{tabular}{lccccc}
\hline
\multirow{2}{*}{\textbf{Speech Encoder}} &\multirow{2}{*}{\textbf{\#Encoder Params}} &\multirow{2}{*}{\textbf{Hidden Size}} & \multirow{2}{*}{\textbf{\#Projector Params}} & \multicolumn{2}{c}{\textbf{WER(\%) $\downarrow$}} \\
& & & & \textbf{test-clean} & \textbf{test-other} \\
\hline
\multicolumn{6}{l}{\textbf{\textit{Acoustic Feature}}} \\
\hline
FBank & - & 80 & 10.03M & 68.95 & 99.37 \\
\hline
\multicolumn{6}{l}{\textbf{\textit{Supervised Speech Encoder}}} \\
\hline
Whisper-tiny & 7.63M & 394 & 12.33M & 7.07 & 16.01 \\
Whisper-base & 19.82M & 512 & 13.64M & 5.07 & 13.07 \\
Whisper-small & 87.00M & 768 & 16.26M & 4.19 & 9.50 \\
Whisper-medium & 305.68M & 1024 & 18.88M & 2.72 & 6.79 \\
Whisper-large & 634.86M & 1280 & 21.50M & 2.58 & 6.47 \\
\quad + Qwen-Audio Fine-tuning & 634.86M & 1280 & 21.50M & 2.52 & 6.35 \\
\hline
\multicolumn{6}{l}{\textbf{\textit{Self-supervised Speech Encoder}}} \\
\hline
HuBERT Base & 94.70M & 768 & 16.26M & 5.39 & 11.99 \\
WavLM Base & 94.38M & 768 & 16.26M & 4.14 & 9.66 \\
HuBERT Large & 316.61M & 1024 & 18.88M &  4.53 & 8.74 \\
\quad + LS-960 Fine-tuning & 316.61M & 1024 & 18.88M &  2.30 & 4.53 \\
WavLM Large & 315.45M & 1024 & 18.88M & 2.37 & 4.90 \\
HuBERT X-Large & 964.32M & 1280 & 21.50M & 4.29 & 6.66 \\
\quad + LS-960 Fine-tuning (SLAM-ASR) & 964.32M & 1280 & 21.50M & 1.94 & 3.81 \\
\hline
\end{tabular}
}
\end{table*}

\begin{table*}[htbp]
\centering
\caption{Compared with other LLM-based speech models. The specific information of the different modules is given in the table.}
\label{tab:slam_asr_llm}
\resizebox{1\linewidth}{!}{
\begin{tabular}{lccccccccc}
\hline
\multirow{2}{*}{\textbf{Model}} &\multicolumn{2}{c}{\textbf{Speech Encoder}} &\multicolumn{2}{c}{\textbf{LLM}} &\multicolumn{2}{c}{\textbf{Projector}} &\multirow{2}{*}{\textbf{ASR Data(h)}} &\multicolumn{2}{c}{\textbf{WER(\%) $\downarrow$}}\\
& \textbf{Module} & \textbf{Learnable} & \textbf{Module} & \textbf{Learnable} & \textbf{Module} & \textbf{Learnable} & & \textbf{test-clean} & \textbf{test-other} \\
\hline
\multicolumn{10}{l}{\textbf{\textit{LLM-based ASR-specific Models}}} \\
\hline
\multirow{2}{*}{\citeposs{yu2023connecting}} & \multirow{2}{*}{Whisper-large} & \multirow{2}{*}{\CROSS} & \multirow{2}{*}{Vicuna-13B} & \multirow{2}{*}{\CROSS} & \multirow{2}{*}{seg-QF} & \multirow{2}{*}{\CHECK} & 960 & 2.3 & 5.2 \\
&  &  &  & & & & 4,000+ & 2.1 & 5.0 \\
\hdashline
\textbf{SLAM-ASR} & HuBERT X-Large & \CROSS & Vicuna-7B & \CROSS & Linear & \CHECK & 960 & \textbf{1.9} & \textbf{3.8} \\
\hline
\multicolumn{10}{l}{\textbf{\textit{LLM-based Audio-universal Models}}} \\
\hline
SALMONN~\cite{tang2023salmonn} & Whisper-large, BEATs & \CROSS & Vicuna-13B & LoRA & win-QF & \CHECK & 1960 & 2.1 & 4.9 \\
Qwen-Audio~\cite{chu2023qwen} & Whisper-large & \CHECK & Qwen-7B & \CROSS & Linear & \CHECK & 30,000+ & 2.0 & 4.2 \\
\hline
\end{tabular}
}
\end{table*}

\subsection{Exploration on Speech Encoders}
\label{sec:encoder}
Furthermore, we fix Vicuna as the LLM and benchmark the performance of different speech encoders. 
For the supervised speech encoders, the performance gets better gradually as the parameter size of the speech encoder increases, which is consistent with the trend on the LLaMA series models. 
When the Qwen-Audio Encoder is used as the speech encoder, the ASR performance is further improved compared with Whisper-large, which indicates that the encoder fine-tuned on other LLM (i.e. Qwen-7B) with gradient backpropagation, can be transferred to another LLM (i.e. Vicuna-7B), and maintain a certain degree of performance. 

For the self-supervised learning speech encoders, HuBERT Base and WavLM Base have about 95M parameters, with 768 dimensions of hidden size.
In this configuration, the ASR performance is similar compared with Whisper-small with the same scale, where self-supervised learning does not play a role. 
When scaling the self-supervised speech encoders to 0.3B, WavLM Large outperforms all listed supervised speech encoders, including Whisper-medium with 0.3B parameters and Whisper-large with 0.6B parameters, while the improvement from HuBERT Base to HuBERT Large is not obvious. 
However, if the HuBERT Large encoder is first fine-tuned on Librispeech 960 hours of training data, and used as the speech encoder to train the projector in our LLM-based ASR model, the model achieves a WER of $2.30\%$ on test-clean and $4.53\%$ on test-other, exceeding the performance with WavLM Large as the speech encoder. 
Further, we use HuBERT X-Large as the speech encoder, which scales the speech encoder to 1B parameters. 
With Librispeech-960 fine-tuned HuBERT X-Large, our LLM-based ASR model gets a word error rate of $1.94\%$ on test-clean and $3.81\%$ on test-other, achieving $24.8\%$ and $41.1\%$ relative WER reduction over the model with Whisper-large as the speech encoder, respectively. 
Additionally, inspired by Fuyu~\cite{fuyu}, we also try to drop the speech encoder and directly feed the 80-dimensional FBank features into the projector, which lags far behind utilizing well-trained speech encoders, as shown in the first row of Table~\ref{tab:benchmark_encoder}. 
The experimental results show the effectiveness of using self-supervised speech encoders and scaling the size of speech encoders. 

\begin{table}[htbp]
\centering
\caption{Compared with previous NN-based models. \textbf{\textit{Specialist Models}} means models trained on Librispeech-960, and \textbf{\textit{in-domain LM}} means language models trained on the LibriSpeech language model corpus along with LibriSpeech-960 transcripts. \textbf{\textit{Universal Models}} means general-propose models trained on massive pair data.}
\label{tab:slam_asr_nn}
\resizebox{1\linewidth}{!}{
\begin{tabular}{lcc}
\hline
\multirow{2}{*}{\textbf{Model}} & \multicolumn{2}{c}{\textbf{WER(\%) $\downarrow$}} \\
& \textbf{test-clean} & \textbf{test-other} \\
\hline
\multicolumn{3}{l}{\textbf{\textit{Specialist Models}}} \\
\hline
ContextNet-large~\cite{han2020contextnet} & 2.1 & 4.6 \\
\quad + in-domain LM & 1.9 & 4.1 \\
Conformer-large~\cite{gulati2020conformer} & 2.1 & 4.3 \\
\quad + in-domain LM & 1.9 & 3.9 \\
Branchformer-large~\cite{peng2022branchformer} & 2.4 & 5.5 \\
\quad + in-domain LM & 2.1 & 4.5 \\
Zipformer-large~\cite{yao2023zipformer} & 2.0 & 4.4 \\
\quad + in-domain LM & 1.9 & 3.9 \\
\hline
\multicolumn{3}{l}{\textbf{\textit{Universal Models}}} \\
\hline
Whisper-large-v2~\cite{radford2023robust} & 2.7 & 5.2 \\
OWSM-v3.1~\cite{peng2024owsm} & 2.4 & 5.0 \\
\hline
\multicolumn{3}{l}{\textbf{\textit{Ours}}} \\
\hline
SLAM-ASR & 1.9 & 3.8 \\
\hline
\end{tabular}
}
\end{table}
\vspace{-0.3cm}

\subsection{SLAM-ASR}
Here we introduce SLAM-ASR, a llm-based ASR model with HuBERT X-Large as the speech encoder and Vicuna-7B as the LLM, with the only trainable linear projector, implemented based on the SLAM-LLM framework. 
As shown in Table~\ref{tab:slam_asr_llm}, we exhibit different LLM-based ASR models from concurrent work, either ASR-specific or audio-universal. 
A contemporary work~\cite{yu2023connecting} employs Whisper-large as the speech encoder and Vicuna-13B as the LLM. The segment-level Q-Former (seg-QF) is utilized as the projector to tackle the compatibility between speech sequences and the LLM. 
Compared with their method, our SLAM-ASR yields $17.4/26.9\%$ relative WER reductions on test-clean/other subsets trained with the same $960$ hours of Librispeech data. 
When their model is trained on a larger amount of speech over $4,000$ hours, the proposed SLAM-ASR still performs better. 
We also compare SLAM-ASR with the latest LLM-based audio-universal models, SALMONN~\cite{tang2023salmonn} and Qwen-Audio~\cite{chu2023qwen}, which provide results on Librispeech benchmark. 
Compared with these audio-based multimodal LLMs, SLAM-ASR still achieves better performance despite the large margin in training data. 

We also compare SLAM-ASR with state-of-the-art previous NN-based models. 
For specialist models trained on Librispeech-960, we compare SLAM-ASR with ContextNet~\cite{han2020contextnet}, Conformer\cite{gulati2020conformer}, Branchformer~\cite{peng2022branchformer}, and Zipformer~\cite{yao2023zipformer}. All models are of large size, and the results from their papers are demonstrated. 
These ASR models employ sophisticated system engineering, including SpecAugment and speed perturbation for data augmentation, and the exponential moving average technique for model averaging. To further improve performance, in-domain language models trained on the LibriSpeech language model corpus along with the LibriSpeech-960 transcripts are added for fusing or rescoring. 
SLAM-ASR achieves the same (test-clean) or better (test-other) ASR performance compared with the best-performing models without using complex system engineering. 
Compared with general-propose models trained on massive data, SLAM-ASR outperforms Whisper-large-v2~\cite{radford2023robust} in industry, and OWSM-v3.1~\cite{peng2024owsm} in the academic community. 
The experimental results demonstrate the superiority of SLAM-ASR and the great potential of LLM-based ASR. 

\section{Capability Emergence}
\vspace{-0.1cm}

We observe that there is capability emergence for LLM-based ASR during training within $1$ epoch (around $12$k steps). 
Specifically, the accuracy of the next token prediction increases rapidly at the beginning of training, then starts to rise slowly, and then ``spikes'' at some point, as if ``the ability is suddenly learned''.

\begin{figure}[htbp]
    \centering
    \includegraphics[width=1\linewidth]{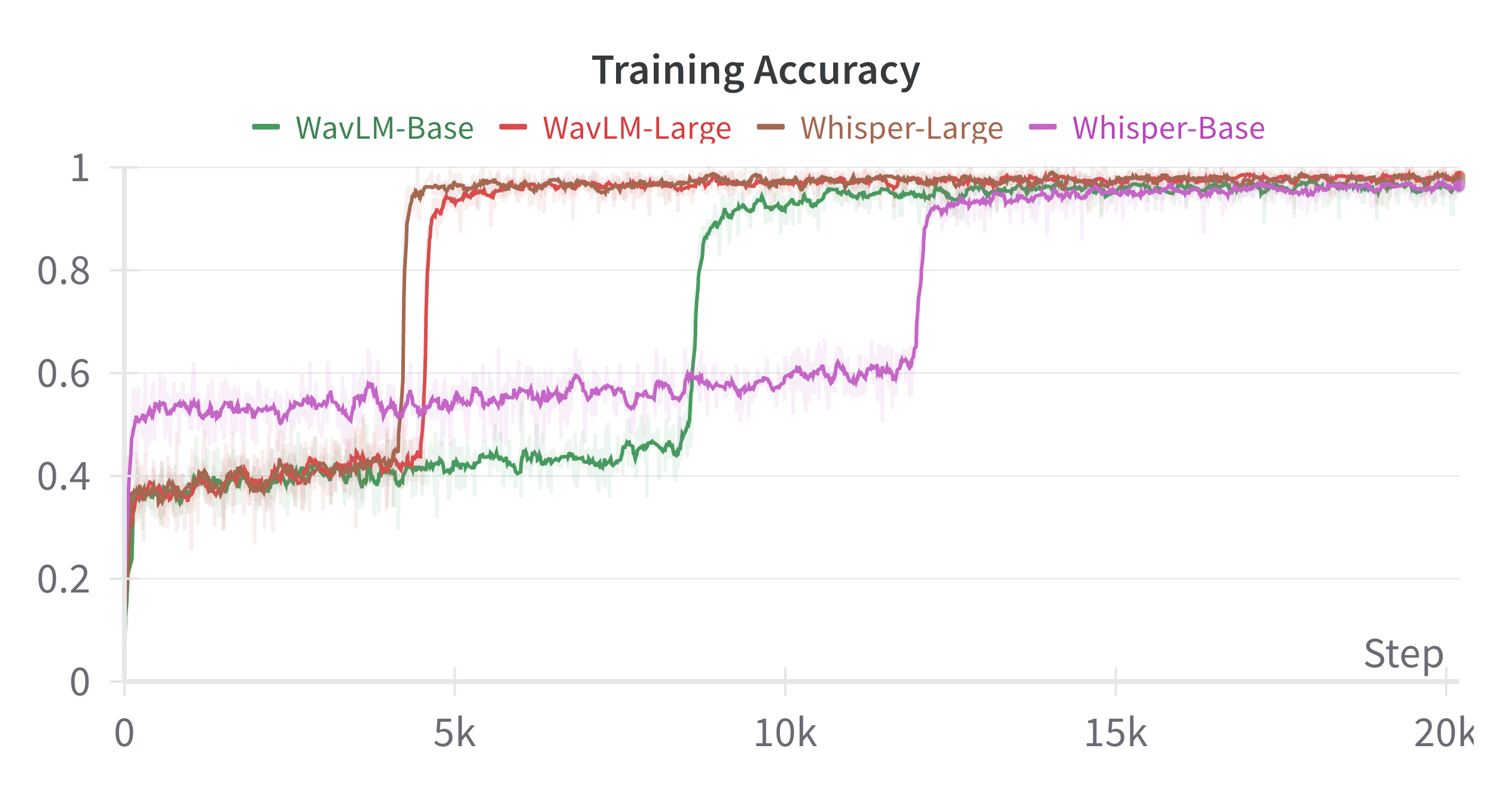}
    \caption{Training accuracy of the next token prediction with the training steps. The LLM is fixed with Vicuna-7B-v1.5 and different colored curves represent different speech encoders.}
    \label{fig:acc_encoder}
    \vspace{-0.3cm}
\end{figure}

Figure~\ref{fig:acc_encoder} demonstrates the training accuracy of the next token prediction with the training steps, where the LLM is kept as Vicuna-7B and the speech encoders vary. 
As can be seen from the figure, the speech encoders with better performance, in this case, Whisper Large and WavLM Large, will emerge earlier. 
A possible explanation is that our task is essentially to align speech representations with LLMs, while a powerful speech encoder can provide representations that are easier for the projector to align with LLMs. 

\begin{figure}[htbp]
    \centering
    \includegraphics[width=1\linewidth]{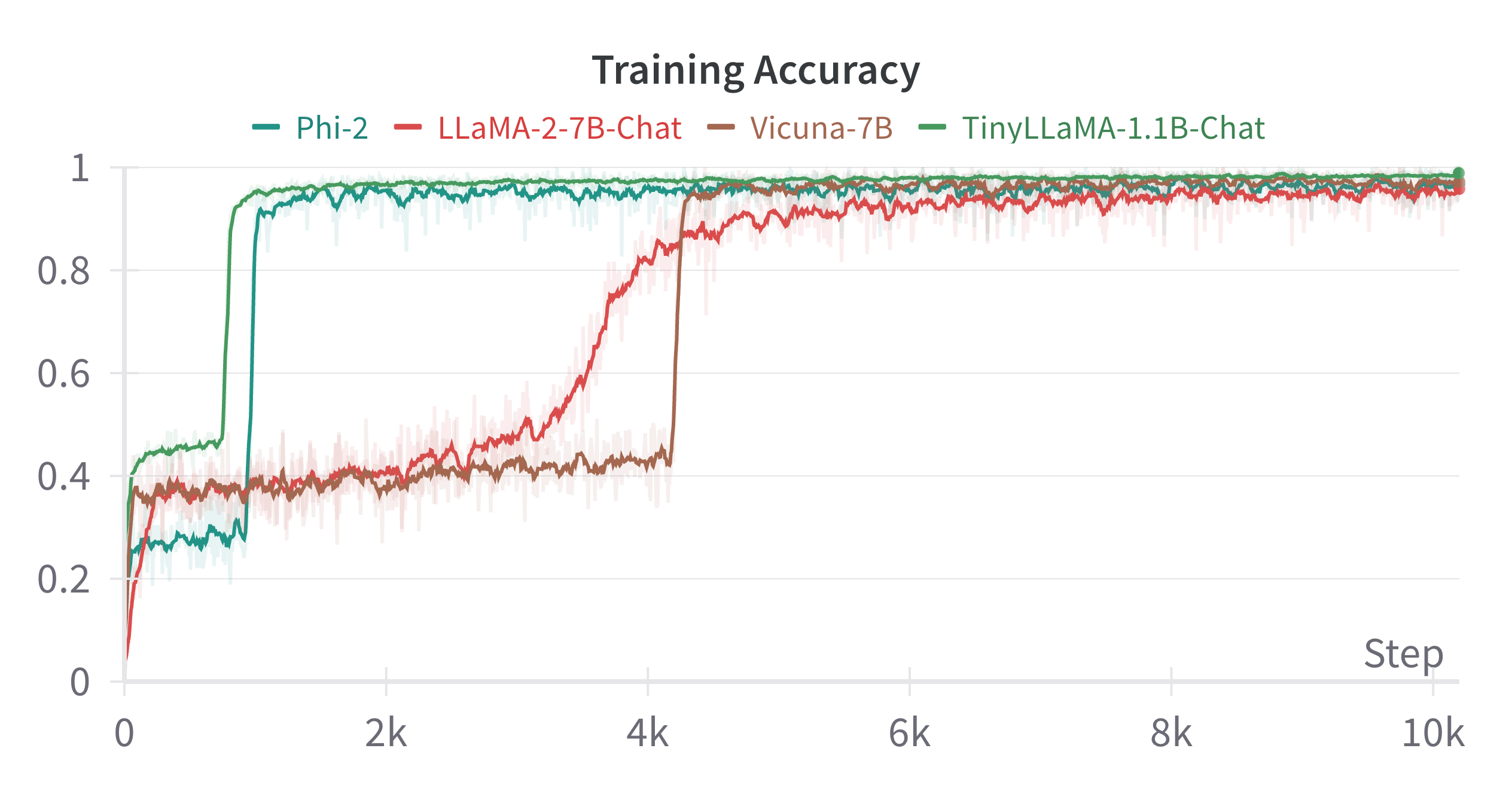}
    \caption{Training accuracy of the next token prediction with the training steps. The speech encoder is fixed with Whisper-large-v2 and different colored curves represent different LLMs.}
    \label{fig:acc_llm}
    \vspace{-0.3cm}
\end{figure}

We keep the speech encoder as Whisper Large, change different LLMs, and plot the training accuracy, as shown in Figure~\ref{fig:acc_llm}. 
Experiments show that LLM-based ASR models with smaller LLMs such as TinyLLaMA-Chat and Phi-2 emerge earlier, however, they are not as effective as larger LLMs such as LLaMA-2-7B-Chat and Vicuna-7B. 
This shows that the larger language models are harder to align with speech features than the smaller ones. 

\begin{figure}[htbp]
    \centering
    \includegraphics[width=1\linewidth]{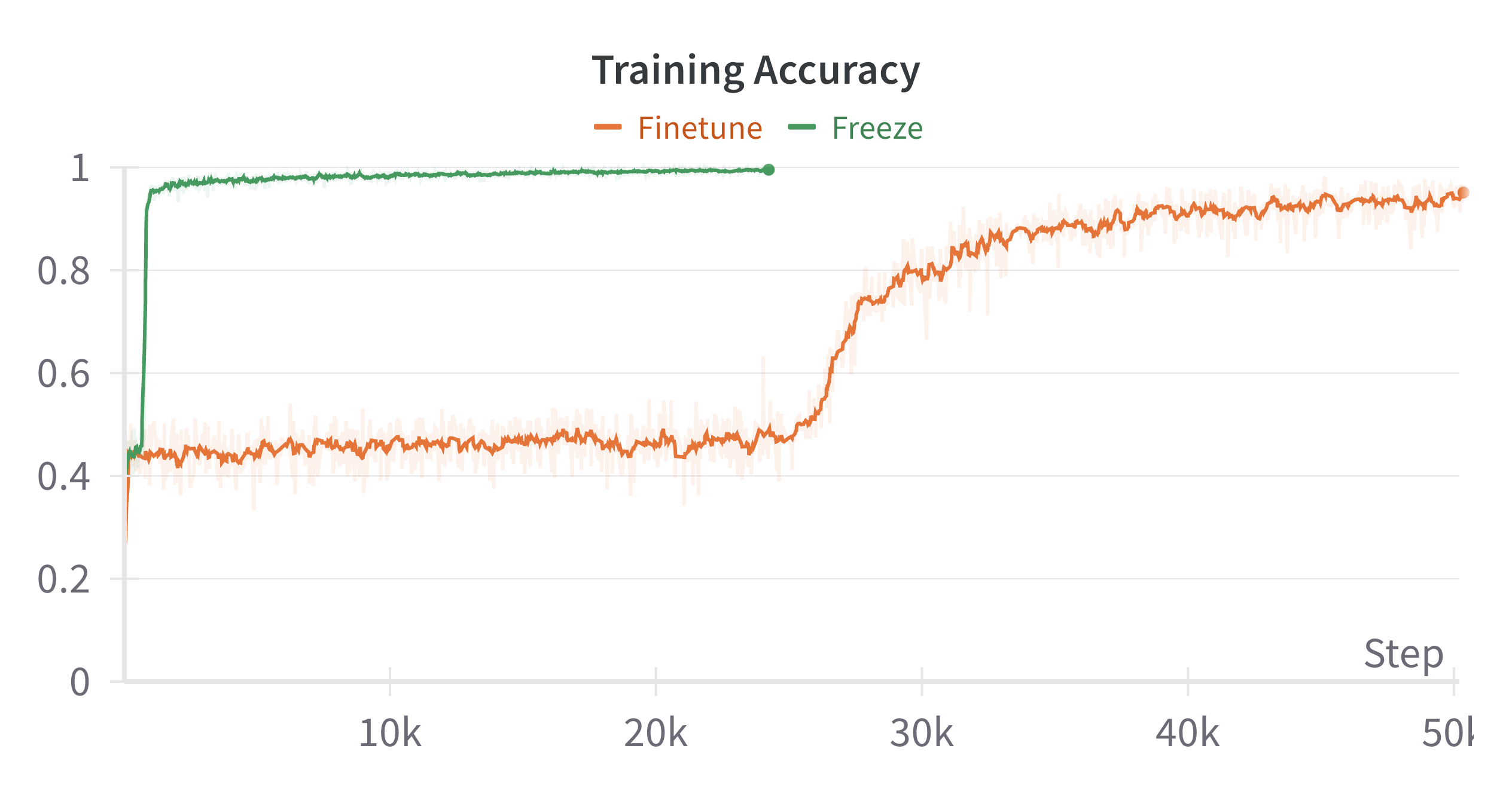}
    \caption{Training accuracy of the next token prediction with the training steps. The speech encoder is fixed with Whisper-large-v2 and the LLM is fixed with TinyLLaMA-1.1B-Chat. Different colored curves represent freezing or fine-tuning the speech encoder. }
    \label{fig:acc_finetune}
    \vspace{-0.3cm}
\end{figure}

We also explore whether or not freezing the speech encoder affects capability emergence. 
We take TinyLLaMA-1.1B-Chat as the LLM and freeze or finetune the speech encoder, respectively. 
As shown in Figure~\ref{fig:acc_finetune}, both models quickly rise to around $40\%$ training accuracy in the early training process. 
When the speech encoder is frozen, the model completes the cross-modal alignment in $1k$ steps, while the time node comes to $25K$ steps when the speech encoder is trainable, which is much later. 
Table~\ref{tab:finetune} compares the WER of the LLM-based ASR systems with the speech encoder freezing and fine-tuning, where the former works much better. 
This indicates that $1k$ hours of speech is still not enough to train a task-specific LLM-based speech encoder, instead, freezing the speech encoder and paying attention to the modal alignment is a better choice. 

\begin{table}[htbp]
\centering
\caption{WER results of freezing or fine-tuning the speech encoder shown in Figure~\ref{fig:acc_finetune} on Librispeech test-clean and test-other subsets. }
\label{tab:finetune}
\begin{tabular}{ccc}
\hline
\textbf{Freezing} & \multicolumn{2}{c}{\textbf{WER(\%) $\downarrow$}} \\
\textbf{Speech Encoder} & \textbf{test-clean} & \textbf{test-other} \\
\hline
\CHECK & 4.33 & 8.62 \\
\CROSS & 12.79 & 22.83 \\
\hline
\end{tabular}
\end{table}

\vspace{-0.4cm}
\section{Conclusion}
In this paper, we systematically explore LLM-based ASR systems with a clean framework, where the only trainable linear projector is used to align the speech encoder and the LLM. 
Research indicates that LLMs that undergo supervised fine-tuning, exhibit improved performance and robustness. 
Furthermore, speech encoders that are fine-tuned from self-supervised models demonstrate superior capabilities. 
The SLAM-ASR model is proposed and outperforms other LLM-based ASR models and previous NN-based ASR models on the Librispeech benchmark. 
Exploratory experiments show that there is a capability emergence in LLM-based ASR systems. 
We aspire for our research to serve as a step forward in the exploration of LLM-based ASR, offering assistance and insights to the broader community. 


\section*{Acknowledgements}
We thank Changli Tang and Wenyi Yu for their helpful discussions and feedback. 

\bibliography{custom}

\newpage
\appendix
\section{Appendix: More Exploration}

\subsection{Text Perplexity}
\begin{table}[htbp]
\centering
\caption{Word-level text perplexity (PPL) and word error rate (WER) of different LLMs on Librispeech test-clean and test-other subsets. Among the listed models, the LLM-based ASR model with Vicuna has the best word error rate, while LLaMA performs the worst. }
\label{tab:ppl}
\resizebox{1\linewidth}{!}{
\begin{tabular}{lcc}
\hline
\multirow{2}{*}{\textbf{LLM}} & \multicolumn{2}{c}{\textbf{PPL (WER(\%)) $\downarrow$}} \\
& \textbf{test-clean} & \textbf{test-other} \\
\hline
LLaMA-2 & 53.74 (3.01) & 58.78 (7.15) \\
LLaMA-2-Chat & 77.60 (2.72) & 85.74 (6.79) \\
Vicuna & 76.44 (2.58) & 84.95 (6.47) \\
\hline
\end{tabular}
}
\end{table}
Word-level text perplexity (PPL) of different LLMs is measured to investigate if the better performance of Vicuna is related to domain agreement, rather than supervised fine-tuning. 
As shown in Table~\ref{tab:ppl}, we measure perplexity on test-clean and test-other subsets. Surprisingly, LLaMA-2 without SFT achieves the lowest perplexity by a large margin compared with chat models, while performing the worst on the word error rate. 
This proves that the better results of chat models are not due to domain agreement with the transcripts. 

\subsection{Prompt Engineering}

\begin{table}[ht]
    \centering
    \caption{Examples of prompts in LLM-based ASR.}
    \label{tab:prmopt_example}
    \resizebox{1\linewidth}{!}{
    \begin{tabular}{ll}
    \toprule
       \textbf{Type} & \textbf{Example} \\
    \midrule
        \textbf{short prompts} & Transcribe speech to text.  \\
    \midrule
        \multirow{4}{*}{\textbf{long prompts}} & Transcribe speech to text. Output \\ 
        & the transcription directly without \\
        & redundant content. Ensure that the   \\
        & output is not duplicated.   \\
    \bottomrule
    \end{tabular}
    }
\end{table}

We also investigate the performance of different prompts in LLM-based ASR, and the prompt examples are shown in Table~\ref{tab:prmopt_example}. 
As shown in Table~\ref{tab:prompt}, when we use a short prompt, the model achieves better results compared with the model using a long prompt in a complex description. However, when we don't use any prompt (that is, a shorter prompt only with the ``ASSISTANT'' tag left), the performance of the model drops. 
This indicates that although an LLM-based ASR model is a task-specific MLLM, the setting of the prompt is still important. 
A possible explanation is that the prompt lets the model optimize in the task-specific subspace through in-context learning, while too complex prompts will increase the learning difficulty and lead to a suboptimal solution.
To investigate this assumption, we set a more complex prompt format. 
We use the same seed prompt for the ASR task in SpeechGPT~\cite{zhang2023speechgpt} to generate $10$ prompts to form a prompt library. At both the training and testing stages, a random prompt is drawn from the prompt library. 
As shown in the last row of Table~\ref{tab:prompt}, there is a big drop in model performance, which is in line with our assumption. 

\begin{table}[htbp]
\centering
\caption{The performance with different prompt designs in LLM-based ASR on
Librispeech test-clean and test-other subsets.}
\label{tab:prompt}
\resizebox{1\linewidth}{!}{
\begin{tabular}{lcc}
\hline
\multirow{2}{*}{\textbf{Prompt}} & \multicolumn{2}{c}{\textbf{WER(\%) $\downarrow$}} \\
& \textbf{test-clean} & \textbf{test-other} \\
\hline
no prompts & 3.19 & 6.97 \\
short prompts & 2.58 & 6.47 \\
long prompts & 2.88 & 6.79 \\
randomly selected prompts & 5.90 & 10.02 \\
\hline
\end{tabular}
}
\end{table}

\end{document}